\documentclass{article} 

\usepackage{arxiv}

\usepackage{microtype}
\usepackage{graphicx}
\usepackage{subfigure}
\usepackage{authblk}
\usepackage{booktabs} 
\usepackage{amsmath,amsthm}
\usepackage{hyperref}

\usepackage{bbm}

\newcommand{\Com}{\eta}

\newcommand{\con}{\mathbf{c}}

\newcommand{\x}{\mathbf{x}}

\newcommand{\s}{\mathbf{s}}
\newcommand{\E}{\mathbb{E}}
\newcommand{\p}{f}

\newcommand{\f}{\phi}

\newcommand{\R}{\mathbb{R}}
\theoremstyle{definition}
\newtheorem{example}{Example}[section]
\newcommand*{\medcup}{\mathbin{\scalebox{1.5}{\ensuremath{\cup}}}}%

\newlength\myindent
\usepackage{amsmath}
\usepackage{amssymb}
\usepackage{mathtools}
\usepackage{amsthm}

\usepackage[capitalize,noabbrev]{cleveref}

\theoremstyle{plain}

\theoremstyle{definition}

\theoremstyle{remark}

\usepackage[textsize=tiny]{todonotes}

\begin{document}

\title{Human-Centered Concept Explanations \\ for Neural Networks}
\date{}
\author[1]{Chih-Kuan Yeh}
\author[2]{Been Kim}
\author[1]{Pradeep Ravikumar}
\affil[1]{Machine Learning Department, Carnegie Mellon University}
\affil[2]{Google Brain}



\maketitle
\begin{abstract}
    Understanding complex machine learning models such as deep neural networks with explanations is crucial in various applications. Many explanations stem from the model perspective, and may not necessarily effectively communicate why the model is making its predictions at the right level of abstraction. For example, providing importance weights to individual pixels in an image can only express which parts of that particular image are important to the model, but humans may prefer an explanation which explains the prediction by concept-based thinking. In this work, we review the emerging area of concept based explanations. We start by introducing concept explanations including the class of Concept Activation Vectors (CAV) which characterize concepts using vectors in appropriate spaces of neural activations, and discuss different properties of useful concepts, and approaches to measure the usefulness of concept vectors. We then discuss approaches to automatically extract concepts, and approaches to address some of their caveats. Finally, we discuss some case studies that showcase the utility of such concept-based explanations in synthetic settings and real world applications.\footnote{This is a book chapter, and the definitive, peer reviewed and edited version of this article is published in [Neuro-Symbolic Artificial Intelligence: The State of the Art, volume: 342, 337 - 352, 2022, https://ebooks.iospress.nl/DOI/10.3233/FAIA210362].}
\end{abstract}

\section{Introduction}
\vspace{-3mm}

Deep neural networks have been instrumental in many modern Artificial Intelligence (AI) successes, with super-human performance across application areas such as vision recognition, speech recognition, and language understanding \cite{he2016deep, hinton2012deep, vaswani2017attention}. As the performance and complexity of deep neural networks improve, understanding how they operate has also become increasingly difficult. This has led to the burgeoning research area of explainable AI (XAI), which provides tools that enable us to better understand an AI model. 

The initial set of XAI methods have focused on providing importance weights for either input  features~\cite{selvaraju2016grad, ribeiro2016should, sundararajan2017axiomatic,yeh2019on}, or for training samples~\cite{koh2017understanding, khanna2019interpreting, yeh2018representer}. While useful, these explanations may not necessarily effectively communicate why the model is making its predictions \emph{at the right level of abstraction}. For example, provides importance weights to individual pixels in an image can only express which parts of that particular image are important to the model. However, human reasoning often comprises ``concept-based thinking,'' where we can loosely relate concepts to groupings of ``similar'' examples~\cite{ARMSTRONG1983263, tenenbaum1999bayesian, margolis1999concepts, rosch1999reclaiming}. Thus, we may obtain better human-centered explanations via \emph{concept-based explanations}, where we explain AI models using human-centered ``concepts''.

In this work, we review the emerging area of concept based explanations. We start by reviewing common classes of explanations, as well as so-called self-interpretable models that use latent variables as concepts. We then review the class of Concept Activation Vectors (CAV) which characterize concepts using vectors in appropriate spaces of neural activations. We next discuss properties of useful concepts, and approaches to measure the usefulness of concept vectors. We then discuss approaches to automatically extract concepts, and approaches to address some of their caveats. Finally, we discuss some case studies that showcase the utility of such concept-based explanations.

\section{Related Work}

While this work is mainly focused on human-centered concept explanations, we start with a brief overview of other classes of explanations. 
%
%
As we will discuss in the sequel, these alternative explanation approaches can in turn be used as a sub-routine within concept based explanations, for instance to explain the importance of individual concepts to  the final model output. In Sec.~\ref{chihkuanyehsec:latent}, we discuss existing work on approaches that use ``semantically meaningful'' latent variables. These extracted latent variables are similar in spirit to concepts, but with the contrast that they are connected to bespoke generative models or specially designed neural networks, whereas concept based explanations~\cite{kim2018interpretability} could in principle be provided for general neural network models.


Certain models are inherently interpretable e.g. small decision trees, but for others e.g. deep neural networks, we need to provide post-hoc explanations, which can in principle be applied to any given (or specific classes of) pre-trained models. Most post-hoc explanations fall under the following three categories (note that the categories are not necessarily disjoint): (a) feature-based explanation methods, that attribute the model output to input features, (b) example-based explanation methods, that attribute the model output to training samples, and (c) counterfactual explanations, which answer ``what if'' questions posed to the model.

\subsection{Feature based Explanations}
Feature based explanations attribute the model prediction to individual features. A large class of feature-based explanations are based on feature perturbations, and gauge the importance of features by perturbing them and measuring the prediction difference. In this line of work,  \cite{zeiler2014visualizing, fong2017interpretable} use perturbations with grey patch occlusions on CNNs, while \cite{DBLP:conf/iclr/ZintgrafCAW17, chang2018explaining, fong2019understanding, petsiuk2018rise, dabkowski2017real, fong2019explanations} improve upon these perturbations via generative models and advanced smoothing designs. Gradient-based explanations can also been seen as a variant of perturbation based explanations via infinitesimal perturbations ~\cite{baehrens2010explain,simonyan2013deep,springenberg2014striving, selvaraju2016grad}, and which range from explicit gradients, to variants that modify back-propagation (such as ignoring negative gradients) to address some caveats with simple gradients. As shown in~\cite{ancona2017unified}, many recent explanations such as $\epsilon$-LRP~\cite{bach2015pixel}, Deep LIFT~\cite{shrikumar2017learning}, and Integrated Gradients~\cite{sundararajan2017axiomatic} can also be seen as variants of gradient explanations.
To reduce the noise in gradient saliency maps, \cite{kindermans2017patternnet} propose removing distractors from the image. SmoothGrad~\cite{smilkov2017smoothgrad} on the other hand generates artificially noisy images via additive Gaussian noise, and averages the gradient of the sampled images; due to the averaging, the resulting output is much less sensitive than vanilla gradients.  \cite{ribeiro2016should} consider a local region characterized by local perturbations of the test input, and restricted to this region approximate the behavior of the given complex model by a locally linear interpretable model. This has been further specialized to different domains by \cite{ying2019gnn, plumb2018model}. \cite{datta2016algorithmic, lundberg2017unified} further compute the importance of a feature by taking the marginal contribution of the feature to any given subset of the set of all features, and computing a weighted average over all possible  subsets; which has roots in cooperative game theory and revenue division. 

Any such feature explanation in turn can be related to concept based explanations by simply treating the set of concepts as features to the model, and obtaining the importance weights of the concept-features. We discuss this further in Sec. \ref{chihkuanyehsec:score}


\subsection{Sample based Explanations}
Sample based explanations attribute the model prediction to individual training samples. Prototype selection methods \cite{bien2011prototype,kim2014bayesian} explain a model by providing a set of ``representative'' samples from the training data set. \cite{kim2016examples} additionally provide criticism alongside the  prototypes to explain what are not captured by prototypes. \cite{koh2017understanding} provide tractable approximations to estimating \emph{influential} training samples defined as those training examples that are the most helpful for reducing the model prediction loss. This was further accelerated by \cite{guo2020fastif} by using a k-nearest-neighbors based selection over training samples. \cite{anirudh2017influential} use a graph over the training samples to select influential training samples. 
\cite{khanna2019interpreting} use the Fisher kernel to select important training examples around the model's decision boundary.
\cite{yeh2018representer} use a representer theorem \cite{scholkopf2001generalized} for neural networks to decompose the neural network prediction as a sum of weighted kernel similarities between the test point and the training samples, which they then use to derive influential training samples. \cite{pruthi2020estimating} propose another decomposition but of the loss of the model at the test point based on the NTK kernel between training points and the test point for different weights along the training trajectory.

Any such sample explanation in turn can be related to concept based explanations by associating clusters of the important training examples with concepts~\cite{ghorbani2019automating}. 

\subsection{Counterfactual and Contrastive Explanations} 
Counterfactual and contrastive explanations \cite{dhurandhar2018explanations,Hendricks2018GroundingVE,vanderwaa2018,goyal2019counterfactual,Joshi2019Towards, Poyiadzi2020FACE, hsieh2020evaluations} answer the question of what to alter in the current input to change the model outcome. Such a contrastive perspective is very amenable to interactive explanations that enable users to understand the model~\cite{Joshi2019Towards,Poyiadzi2020FACE}. Such counterfactual explanations can be seen to be related to adversarial examples~\cite{carlini2017towards, goodfellow2014explaining} as they both try to find small perturbations to the test input that changes the model prediction~\cite{Wachter2017Counterfactual}. \cite{xu2018structured} add group sparsity regularization to improve the semantic structure of such adversarial perturbations. \cite{ribeiro2018anchors} deems a set of features important if the model prediction does not change a lot when only perturbing features not in the set. In related work, \cite{hsieh2020evaluations} deem a set of features as important if perturbations within the set of features changes the model outcome, while perturbations outside the set do not. 

Any such counterfactual and contrastive explanation in turn can be related to concept based explanations by asking the question ``what concepts to alter to change the outcome of the model''. This is explored by a line of recent works \cite{koh2020concept, goyal2019explaining, sreedharan2020bridging, ghandeharioun2021dissect}, and which we  discuss further in Sec. \ref{chihkuanyehsec:concept}.

\subsection{Semantically Meaningful Latent Variables and Self-Interpretable Concept Networks} \label{chihkuanyehsec:latent}

A line of work focuses on deriving semantically meaningful latent variables or features, which can naturally be related to concepts. A popular class of these approaches use dimensionality reduction methods (such as variational encoders \cite{kingma2013auto}) to derive latent features that can be connected to human-relatable concepts~\cite{chan2015pcanet, kingma2013auto}, and which has shown great success in applications such as speech~\cite{chorowski2019unsupervised} and language~\cite{unsupervised_sentiment, 8668423}. \cite{locatello2018challenging} however, show that meaningful latent features cannot be derived in a completely unsupervised setting, and suggest the necessity of appropriate inductive biases.

Another line of work focuses on designing ``self-interpretable'' models that themselves have semantically-meaningful latent variables. 
\cite{chen2019looks} develop such a self-interpretable model for image classification by relating representative training patches to latent variables, while \cite{shi2020corpus} develop ones for language data.
\cite{bouchacourt2019educe} further use ``concept networks'' in their self-interpretable models. \cite{alvarez2018towards} extend such designs to improve their robustness. \cite{chen2020concept} propose to replace batch normalization in deep neural networks with a concept whitening module. The benefit of such self-interpretable models is that they do not make the common assumption that the concept vectors lie in a linear vector space of some neural network layer activations (which we will expand upon in the sequel). The caveat with these models on the other hand is that they are not applicable as a post-hoc explanation approach for arbitrary models.

\section{Concept based Explanations: A Human-centered Approach} \label{chihkuanyehsec:concept}

Suppose we denote the given model by $\p: \R^d \mapsto \R^K$, so that given a test input $\x \in \R^d$, it outputs $\p(\x) \in \R^K$ as the $K$ scores for each of $K$ classes. The class-specific output for the $k$-th class is then given by $\p_k(\x) \in \R$. We additionally use $\p^{[\ell]}(\x) \in \R^{d^{[\ell]}}$ to denote the $l$-th layer activation for test point $\x$. A concept can then be loosely defined as a vector that ``represents'' the collection of activation vectors (for the given model) corresponding to concept relevant inputs. In the next couple of sections, we discuss approaches that formalize this further.


\subsection{CAV and TCAV}


\cite{kim2018interpretability} introduce Concept Activation Vectors (CAV) motivated by a line of research that the linear vector space of neural activations may encode meaningful, insightful information~\cite{mikolov2013efficient, pennington2014glove}. They define CAV as the normal vector to a hyperplane separating non-concept-example activations from concept-example activations. Such a hyperplane can be obtained by a binary classifier that discriminates between positive and negative examples with respect a given concept. More formally, suppose we are given set $P$ of positive examples and a set $N$ of negative examples with respect a given concept, for a total of $n$ examples. Then the intermediate activations of positive examples is given by $\{\p^{[\ell]}(\x): \x \in P\}$, while those for negative examples is given by $\{\p^{[\ell]}(\x): \x \in N\}$.
The concept activation vector $c$ for the given concept can then be obtained as: \[c = \arg\min_{c'} \min_{b} \sum_{j=1}^{n} L(y_j, \p^{[\ell]}(\x_j) \cdot c + b),\] where $y_j$ = 1 if $\x_j \in P$ and $y_j$ = 0 if $\x_j \in N$, and $L$ is some loss function for binary classification. The concept vector thus acts as a linear discriminant in the activation vector space between positive and negative examples for a concept.

To quantify how relevant a given concept is to the prediction of the model, \cite{kim2018interpretability} propose the TCAV score, which is the fraction of training samples whose model classifier scores increase when the input is infinitesimally moved in the direction of the concept. To state this formally, for a concept $c$, class label $k$, and layer $\ell$, and using $\mathcal{X}_k$ to denote the set of training samples with class $k$, we have: \[TCAV_{c,k,l } = \left|\left\{\x\in \mathcal{X}_k:  \frac{d\p_k(\x)}{d\p^{[\ell]}(\x)}\cdot c >0\right\}\right|/|\mathcal{X}_k|.\]
A concept is considered to be related to a class label $k$ if the TCAV score is significantly different from TCAV scores with random concepts (i.e., where concept examples are random examples).

Given \emph{relative concepts} such finer-grained comparisons (e.g., brown hair vs. black hair), we can then obtain relative CAVs, and relative TCAV scores. These are most useful in determining which among set of concepts is most relevant to the model, in contrast to determining whether a concept is relevant to the model at all~\cite{cai19}. 

\subsection{Interpretable Concept Basis}

Similar to CAV, \cite{zhou2018interpretable} define an \emph{interpretable concept basis} based on the linear discriminant on an intermediate layer $\p^{[\ell]}$, obtaining concept vectors $c = \arg\min_{c'} \min_{b} \sum_{j=1}^{n} L(y_j, \p^{[\ell]}(\x_j) \cdot c + b)$,
given binary concept labels $\{y_j\}_{j=1}^{n}$ for a set of inputs $\{\x_j\}_{j=1}^{n}$; following the same notation as in the earlier section. Here, they specify the intermediate layer $\p^{[\ell]}$ to be the second to last layer just before the logit scores, which are thus given by:
\[\p_{\text{logit}}(\x)=  W \p^{[\ell]}(\x) + b,\]
for a weight matrix $W$ and bias vector $b$,
The prediction logit for class k is then given by:
\[\p_{\text{logit};k}(\x)= w_k \cdot \p^{[\ell]}(\x) + b_k.\]

The goal of \cite{zhou2018interpretable} is to then decompose the weight vectors $w_k$ in terms of the concept vectors $\{c_{j}\}_{j=1}^{m}$, such that 
\[w_k \approx \alpha_{k1} c_1 + \alpha_{k2} c_2 + ... + \alpha_{km} c_m.\]

The weights $\alpha_k := \{\alpha_{kj}\}_{j=1}^{m}$ are further constrained to be non-negative since negative weights are harder to interpret; and are further constrained to have sparsity $s < m$. To address these desiderata, they propose to solve for weights $\alpha_k := \{\alpha_{kj}\}_{j=1}^{m}$ via the following optimization problem:
\[\arg \min_{\alpha_k >0, |\alpha_k|_0 \leq s} \|w_k - \sum_{j=1}^{m} \alpha_{kj} c_j\|\]


Thus both interpretable concept bases~\cite{zhou2018interpretable} and CAVs~\cite{kim2018interpretability} follow the same recipe for deriving concept vectors. The main distinction between them is that the former learns concepts vectors from the second-to-last layer, and focus on the use of these concept vectors as an interpretable basis for logit weights; while the latter can be applied to any user-chosen layer, and focuses on the use of these concept vectors to determine their importance to the model prediction via TCAV scores.

\subsection{Some Empirical Evidence for Concept-Based Explanations}

\begin{figure}[h] 
\includegraphics[width=0.8 \textwidth]{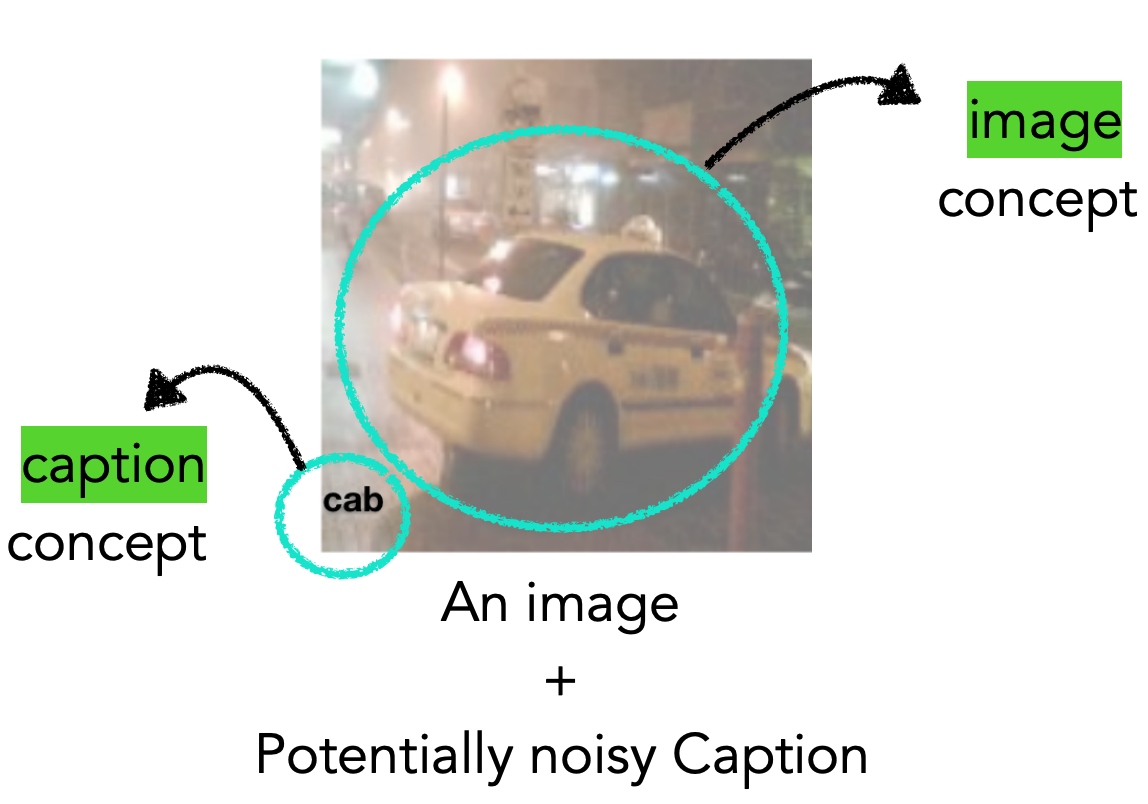}
\caption{Example image containing both the image of a cab and the text caption of cab.}\label{chihkuanyehfig:example_sanity}
\end{figure}

\cite{kim2018interpretability} propose an experiment based on a dataset of images, that have objects, such as cucumbers and cabs, along with a text caption, such as ``cucumber'' and ``cab,'' on the bottom of the image. By varying the noise in the image and the text caption, they obtain several models that each use only the text caption, or only the object in the image for classification. An example image is shown in Fig \ref{chihkuanyehfig:example_sanity}; here the model may classify this image as a car based either on the object in the image, or the text caption. The approximated ground truth of whether the model depends on the image object or text caption is determined by the performance of the model when only the object or text is provided. For instance, if the model performs well only when the image object is provided, but does not perform well when only text is provided, then the model is determined to depend on the image object.

\cite{kim2018interpretability} then conducted a 50-person human study on Amazon Mechanical Turk, and asked users to predict whether the model depends on the text or image object for two models based on saliency map explanations~\cite{sundararajan2017axiomatic, smilkov2017smoothgrad}. Of the two models, one model depends on text, and the other model depends on image object. Only $52 \%$ of users were able to predict the correct model source when given saliency map explanations, while they were almost always able to predict the correct model source when given TCAV score explanations. This is likely due to TCAV showing the importance of human-defined concepts, while saliency scores show the importance of raw features. 

\section{Properties and Evaluation Measures of Concepts}

We next discuss some useful properties of concept vectors, as well as measures to evaluate the usefulness of such concept vectors.

\subsection{When are Concepts Sufficient to Fully Explain the Model?}

A limitation of TCAV scores is that they only answer the question of whether a concept is \emph{related} to the model prediction. However, it may be unclear whether these concepts fully account for the model's prediction. For instance, several concepts may be mildly related to the model prediction, but the main concept critical to the model prediction may not have been discovered. Therefore, it would be useful to determine whether the concepts are also \emph{sufficient} to explain the model prediction. 

In this section, we discuss how to measure the sufficiency of concepts as proposed by \cite{yeh2020completeness}. The high level assumption here is that the concept scores for a sufficient set of concepts 
should be a sufficient statistic for the model output. The intuition being that if this does not hold, the model is likely basing its prediction on information \emph{not captured} by the concepts, and thus the concepts are not ``sufficient'' to explain the model output.

Suppose we are analyzing a set of $m$ concepts, with unit concept vectors $\con := \{\con_j\}_{j=1}^{m}$ that represent directions in the activation vector space $\text{span}(\text{range}(\f))) \subseteq \R^d$. Suppose the overall model prediction can be written as $\p(\x) = h(\phi(\x)))$, where $h(\cdot)$ maps the activations to the model outputs. We also allow for the input $\x = (\x_t)_{t=1}^{T}$ to correspond to multiple parts (e.g. patches in an image, or some specified sets of input features)

We then define the part concept score for the input part $\x_t$ and concept vectors $\con$ as $v_\con(\x_t) := \langle\phi(\x_t),c_j\rangle_{j=1}^{m} \in \mathbb{R}^m$, and the \emph{concept score} for the entire input $\x$ as $ v_\con(\x) = (v_c)_{t=1}^{T} \in \mathbb{R}^{T \cdot m}$. These concept scores thus capture similarities of (activations of) the input (parts) to the concept vectors.

We define ``sufficient'' concepts as those whose concept scores are sufficient statistics for the model output, so that we may evaluate the completeness of concepts by how well one can recover the prediction given the concept scores. Let $g: \mathbb{R}^{T\, m} \rightarrow \mathbb{R}^{T \, d}$ denote any mapping from concept scores to the activation space. If concept scores $v_\con(\cdot)$ are sufficient statistics for the model output, then there exists some mapping $g_\p$ such that $h(g_\p(v_\con(\x))) \approx \p(\x)$. Following this intuition, we define the \textbf{Completeness Score} $\Com_\p(\con)$ for a set of concept vectors $\con := \{\con_j\}_{j=1}^{m}$ as follows:
	    \begin{equation} \label{chihkuanyehseceq:complete}
         \Com_\p(\con_1,..., \con_m) = \frac{\sup_g \mathbb{P}_{\x,y}[{y} = \arg\max_{y'} h_{y'}(g(v_\con(\x)))]-a_r}{\mathbb{P}_{\x,y}[{y} = \arg\max_{y'}\p_{y'}(\x)]-a_r},
	    \end{equation}
where $a_r$ is the accuracy of random prediction to equate the lower bound of completeness score to 0. The population accuracies above can be estimated by the empirical accuracies over validation data. Note that $\sup_g \mathbb{P}_{\x,y \sim V}[{y} = \arg\max_{y'} h_{y'}(g(v_\con(\x)))]$ is the accuracy of the best possible classifier when only given concept scores $v_\con(\x)$. In practice, we optimize over a flexible class of classifiers such as DNNs.

Note that the completeness score can also be used to measure how sufficient concepts are for the dataset itself independent of the model, by replacing $\phi(\cdot), h(\cdot)$ with identity functions, and the model prediction $\p(\x)$ with the label $y$. 
We provide an illustrative example below on the usefulness of the completeness score.
\begin{example} \label{chihkuanyehsecex:com}
Suppose the inputs $\x \in \mathbb{R}^m$, and the activation function  $\phi$ is simply the identity function. Assume that the features $\x_1, \x_2, \hdots, \x_m$ follow independent Bernoulli distributions with bias $p = 0.5$, and the model we attempt to explain is $f(\x) = \x_1 \text{ XOR } \x_2 ... \text{ XOR } \x_m$. A natural set of $m$ concepts $\con_1, \con_2, \hdots, \con_m$ can be obtained by a one-hot encoding of each feature in $\x$. It can be seen that the full set of concepts $\con_1, \con_2, \hdots, \con_m$ are sufficient for the model prediction, so that $\Com_\p(\con_1, \con_2, ..., \con_{m})=1$. However, if we only have information on $\con_1, \con_2, ..., \con_{m-1}$ but not on $\con_m$, then we can only do as well as random guessing in predicting the model output. In this case, $\Com_\p(\con_1, \con_2, \hdots, \con_{m-1})=0$.
\end{example}

The completeness score allows us assess the `sufficiency'' of the discovered concepts to ``explain'' the reasoning behind a model's decision. After users manually label some set of concepts and extract concepts of interest in the TCAV score, one can then calculate the concept completeness score. If the completeness score is too low, users could then try to label additional concepts that might be relevant to the model. 

\subsection{Interventions on Concepts}

Given the development of concept vectors as in the previous sections, a natural follow-up question is ``how would the model prediction change if a concept were changed in some manner in the input?''. 

To address this question, \cite{koh2020concept, goyal2019explaining}  propose frameworks to estimate the effect on model output upon intervening on the concepts, with each making different assumptions on the data generative process. \cite{goyal2019explaining} assume the generative process follows the graphical model $\con \rightarrow \x \rightarrow y$, while \cite{koh2020concept} assume the generative process follows the graphical model $\x \rightarrow \con \rightarrow y$. 
Note that the latter explicitly entails that concepts scores are sufficient for the model prediction, which mirrors the sufficiency and completeness viewpoint concepts proposed by \cite{yeh2020completeness}.

\cite{goyal2019explaining} define the Causal Concept Effect (CaCE) as the causal effect of a binary concept $\con$ on the output of the classifier $\p$ under the generative process above: 
\[\text{CaCE}(\con, \p) = \E[\p(\x|\text{do}(\con = 1)]- \E[\p(\x|\text{do}(\con = 0)].\]


The benefit of the causal concept effect over the TCAV score is that it captures the causal relationship between the concept and model output, while TCAV only captures the associative relationship between the concept and the prediction. Consider the case where the model is solely dependent on the color of the animal (black and white). In the dataset, assume there is an equal number of cat and dogs, but $90\%$ of the cats are white and $90\%$ of the dogs are black. The TCAV model may easily learn that the concept ``cat'' is related to the model output, since the concept vectors of ``cat'' and ``color'' as estimated from this data may be close. However, in spite of this correlation, the causal concept effect may be able to learn that the concept ``cat'' is not causal to the model prediction. To obtain CaCE values, \cite{goyal2019explaining} utilize environments where intervening on concepts is possible, or to approximate these using flexible generative models such as conditional VAE models to generate input conditioning on the concept values. 

\cite{koh2020concept} assume the model prediction score can be computed based solely on the concept scores. They thus train a new model where one of the intermediate layers is exactly the concepts scores for each of the  concepts, and use this new model to make predictions based on the concept scores. The benefit of using this retrained model is that it is then possible to intervene on the concept scores, and answer questions such as ``If the concept score was different for this input, will the model make a correct prediction?''. This also allows user to interact with models and correct certain concept scores for prediction. However, an implicit limitation is that if complete concepts are not provided, it is unclear whether the user would be able to interact with the concept scores as intended.

\subsection{Importance Evaluations of Concepts} \label{chihkuanyehsec:score}

There are many ways to measure the ``importance'' of concepts for a set of concepts $\con_1, \con_2, \hdots, \con_m$ and a model $\p(\cdot)$. Here, we recap a set of quantitative measurements to evaluate the importance of each concept.

\paragraph{TCAV score \cite{kim2018interpretability}:} The TCAV score measures the inner product between the concept vector $\con$ and the gradient of the model prediction for class $k$ with respect to activations in an intermediate layer $\ell$:

\[TCAV_{c,k,l } = \left|\left\{\x\in \mathcal{X}_k:  \frac{d\p_k(\x)}{d\p^{[\ell]}(\x)}\cdot c >0\right\}\right|/|\mathcal{X}_k|.\]

\paragraph{Concept Causal Effect \cite{goyal2019explaining}:} The  concept causal effect of concept $\con$ measures the difference in prediction value of the model $\p$ when intervening on value of a binary concept value $\con$:

\[\text{CaCE}(\con, \p) = \E[\p(\x|\text{do}(\con = 1)]- \E[\p(\x|\text{do}(\con = 0)].\]

\paragraph{ConceptSHAP \cite{yeh2020completeness}:}
Given a set of concepts $C_S = \{\con_1, \con_2, \hdots, \con_{m}\}$, and a completeness score function $\Com$, the ConceptSHAP $\s_i$ for concept $\con_i$ is defined as 
\begin{equation*}
\begin{split}
     \s_i(\Com) = \sum\nolimits_{S \subseteq C_s \backslash\con_i} \frac{(m-|S|-1)!|S|!}{m!}[\Com(S \medcup \{\con_i\}) - \Com(S)],
\end{split}
\end{equation*}

where each concept is treated as a player in a co-operative game. The Shapley value aggregates the contribution of each player based on its performance when co-operating with other players, and uniquely satisfies a natural set of game-theoretic axioms \cite{shapley_1988,lundberg2017unified}.

\paragraph{$\mathbf{TCAV}^\text{ICS}$ \cite{schrouff2021best}:}  \cite{schrouff2021best} introduce $TCAV^\text{ICS}$, which combines TCAV scores with Integrated Gradient (IG) by projecting the IG on the intermediate layer $\p^{[\ell]}(\x)$ onto the concept direction $\con$. The full formulation can be written as:

\[TCAV^\text{ICS}_{\con,k,l, b } = \frac{1}{n}\sum_{i=1}^{n}  (\p^{[\ell]}(\x_i)-b )^T \con \int_{[b,\p^{[\ell]}(\x_i)]} \con \cdot \frac{d\p_k(\x_i)}{d\p^{[\ell]}(\x_i)}  d\p^{[\ell]}(\cdot).\]

Here, $b$ is a baseline value which is set to be the activation after ``removing'' concept information, so that we could set $b = \p^{[\ell]}(\x) - \lambda \con$, or $b = \p^{[\ell]}(\x) - \text{projection}(\p^{[\ell]}(\x), \con)$.

Each of these concept importance scores can be seen to be related to methods of feature importance, where TCAV is related to the model gradient, CaCE is related to Leave-One-Out explanations (measuring effect on model output when leaving out a feature), ConceptSHAP is related to the Shapley value explanations, and $TCAV^\text{ICS}$ is related to the model Integrated Gradient. It would be interesting to see if other approaches for feature importance can be usefully extended to measure the importance of concepts.

\section{Unsupervised Discovery of Concepts}

One of the main caveats for CAVs is that we need to provide training examples for each of the concept. These typically have to be manually labeled, which however may often not be possible or too costly. 

A natural question is whether can one discover concepts in the dataset that are human interpretable in an \emph{unsupervised manner}?  We briefly review a few approaches that address this question. \cite{ghorbani2019towards} discover concepts in images by super-pixel segmentation followed by k-means clustering. \cite{yeh2020completeness} discover concepts in image and language data by learning concepts that are representative of clusters of training inputs. \cite{ghandeharioun2021dissect} find disentangled and diverse concepts (which they term a concept trajectory) by leveraging  counterfactuals.

\cite{ghorbani2019towards} propose the desiderata of ``Meaningfulness'', ``Coherency'', ``Importance'' for concept explanations. To obtain meaningful concepts, they use super-pixel segmentation to mimic the process of humans finding semantically meaningful parts of the image data. To obtain meaningful and coherent parts of the data, they propose to use k-means clustering of the super-pixel segments. Finally, they use TCAV scores to determine if the k-means clusters are related to the model prediction for any specific class.
\cite{yeh2020completeness} aim to find concepts that are ``complete'' with respect to the model, while also ensuring the concepts are coherent and meaningful by: (a) limiting each concept to be a contiguous part of the input data (a patch of image or a sub-sentence in language data), and (b) enforcing each concept to be close to its top-k nearest neighbors in the training data. The first constraint increases the meaningfulness of the concept, as contiguous parts of the data are more interpretable. The second constraints also increase human interpretability of the concepts, as users can better understand the concept by observing the top-k nearest neighbor training examples.
\cite{ghandeharioun2021dissect} aim to find $k$ concepts by generating counterfactuals close to the image of interest $\x$. They use a generative model $G_\x(\alpha,c)$ they call DISSECT that generates counterfactuals of any test image $\x$ by only changing the concept $c$, and where the model prediction on the counterfactual is some predefined value $\alpha$, so that $\p(G_\x(\alpha,c)) = \alpha$. They then use this generative model to learn the concepts themselves. 

A less ambitious approach might be to simply \emph{propose} potential concepts in an unsupervised manner, and which can then be identified and labeled by human experts.\cite{janik2019discovering} provide a visualization of random directions in the activation space in real-time, and which  users can rotate around and label as concepts if they find a direction that is meaningful to them. Another approach is to use statistical test to find interesting directions. While not completely unsupervised, this significantly lowers the labeling cost for concept vectors.

\subsection{Undesired Correlation and Dependence between Concepts and Labels}

A caveat with the approaches above is that they might learn concepts that are predictive of the class label due to so-called spurious associations and correlations between the concepts and the label, but which are not causally related to the label~\cite{mahinpei2021promises, margeloiu2021concept}. A natural reason this might occur is due to what is known as confounding. As an example, consider an MNIST digit classification task of predicting whether a digit is even or odd. Then concept scores for say specific digits of `4' and `5' are predictive of the class label, but are perhaps not the right concepts to be learned. Note that even for a random concept via a random direction in the activation space, the concept scores which are the projection of the input onto the concept direction still encodes predictive information. 
From a generative process viewpoint, this implies that some confounding variables may exist between input data and concepts, and thus concept scores which are not causally related to the model prediction nonetheless have predictive information.


To account for the confounding variables between input data and concepts, \cite{bahadori2020debiasing} propose the following generative model. Let $u$ denote the confounding variable between the input data $\x$ and the concepts $\con$. Let $d$ denote the unconfounded (unbiased) concepts that are independent of the confounding variable $u$. The generative model is then given by:
\begin{align}
    d &= \p_1(y) + \epsilon_1,\\
    c &= d + h(u), \\
    \x &= \p_2(u,d) + \p_3(y) + \epsilon_2,
\end{align}
where $\epsilon_1, \epsilon_2$ are independent noise variables, and $\p_1, \p_2, \p_3$ are deterministic functions. \cite{bahadori2020debiasing} show that debiased concepts from the generative model above can retrieve higher quality concepts in both simulation and real world datasets. Note that even if we use debiased concepts above, they never be fully complete in the sense of \cite{yeh2020completeness}, as $\x$ may contain more predictive information (due to spurious correlations with the label) when compared to the debiased concepts. 
    
 
A related idea to the debiasing of concepts is to ``whiten'' the concepts (so that the covariance of the concepts is the identity matrix). \cite{chen2020concept} suggest using concept whitening modules within a deep neural network, which they suggest learn concepts that do not rely on the usual assumptions that concept vectors lie in the linear vector space of neural activations~\cite{kim2018interpretability}. However, \cite{mahinpei2021promises} show that there is information leakage even when you whiten the concept, since decorrelating concept representations does not remove all statistical dependence. They thus suggest minimizing the mutual information between concept scores to prevent information leakage of soft concept representations.
%
 


 
%
%

\section{Evaluations of Concept Based Explanations}

Given a set of concepts $\con_1, \con_2, \hdots, \con_m$, and a model $\p(\cdot)$, how does one evaluate the quality of the set of discovered concepts? We first discuss the setting where the ground truth concepts are unknown, and then provide some case studies where we do know the underlying concepts.

\subsection{When Ground Truth Concepts Are Unknown}

While most work on concept explanations use specific datasets where ground truth concepts can be retrieved, we next discuss some evaluation measures that can be used even when the ground truth concepts are unknown.

\paragraph{Necessary and Sufficient Concepts \cite{ghorbani2019towards}:} One way to evaluate a set of concepts is to mask out (or only keep) the top $k$ most important concepts in the training data. If the set of concepts is useful for prediction, removing (only keeping) the top-$k$ concepts should result in a large drop (increase) of the model performance. 



\paragraph{Completeness Score \cite{yeh2020completeness}:} The evaluation of the completeness score can also be performed in the absence of knowledge of ground-truth concepts. It can be seen to be similar to the remove and retrain framework above, but where we only partially retrain the part of the model from the intermediate layer activation space to the final prediction. This is because the concept scores of complete concepts are assumed to be a sufficient statistic for the model prediction $\p$.

\paragraph{Synthetic Datasets \cite{yang2019benchmarking}:} \cite{yang2019benchmarking} introduces the BAM dataset, where objects can be synthesized in different scenes. They then train two models, where one is trained with labels of the objects and one is trained with labels of the scene. They then proposes the model contrast scores (MCS), which measure the difference of concept scores of objects in the two models. The concept score of objects should be larger when the object model is used, and the concept score of scenes should be larger when the scene model is used.

One caveat with the first approach to evaluate concepts is that it involves perturbing the input with respect to the concepts, which may take us out of the data manifold. With model contrast scores, the effect of true concepts could be correctly evaluated, but the concept space and the model space would be limited to those in the BAM dataset. Completeness scores can be used without perturbing images and without synthetic data. However, it does not take into account end-task performance.

\subsection{Case Study: Building Post-Hoc Symbolic Explanations for Sequential Decision-Making Problems by User-Defined Concept}

Concept-Based explanations have the benefit of speaking the language of users, and can also be used as a base unit to build symbolic explanations with user-defined concepts. We next describe an example of such a partially-symbolic model based on human-defined concepts~ \cite{sreedharan2020bridging}.
In the game Montezuma’s Revenge, a popular Atari game, the goal is to score points by gathering jewels and killing enemies along the way. Concepts labeled by humans include ``skeleton on left'', ``left door closed'', and ``not on ladder''. A symbolic approximation of an RL agent can then be learned from the concept scores to the model output
with preconditions and cost functions similar to the STRIPS model~\cite{geffner2013concise}. The symbolic approximation model chooses an action if the current state satisfies the preconditions of the action, and if the action has the lowest cost function of all viable actions.

Based on this symbolic approximation model, \cite{sreedharan2020bridging} propose to generate contrastive explanations for why the model is not performing a ``foil action,'' which is an alternate plan specified by users. Based on the symbolic action model, the model presents explanations of the form ``the foil action A is not chosen because precondition B is not satisfied'' or ``the foil action A is not chosen because its predicted cost C is higher than the current plan with cost D''. As an example, the model could present an explanation that the foil action ``go left'' is not chosen because precondition ``skull not on left'' is not satisfied, or that the foil action ``attack'' is not chosen because the cost of attack is at least 500 when ``skull on left'', but the cost of the chosen plan is 20. While this proposed algorithm relies on the assumption that it is possible to approximate the applicability of actions and cost functions in terms of high-level concepts and via a symbolic action model, which may be not satisfied for more complex environments, this nevertheless sheds light on how concept explanations can bridge the gap between black-box model and symbolic reasoning as explanations.

\subsection{Broader Applications of Concept-Explanations}
In addition to extensions in the machine learning community, concept-based explanations have been widely introduced to application areas beyond the machine learning community. In the medical domain, using concept-based explanations \cite{graziani2018regression} show that nuclei texture is a relevant concept in detecting tumor tissue in breast lymph node samples. \cite{clough2019global} show that ventricular ejection and filling rates concepts are crucial in cardiac MRI classification. \cite{yeche2019ubs} discover that radiomic features describing increased entropy, as well as those describing variations of intensity are useful concepts for the prediction of calcifications. In scientific domains, using concept-based explanations, \cite{sprague2018interpretable} show that the ``Eye'' concept is important for the prediction of Category 4 tropical cyclones in a CNN model. \cite{mincu2021concept} show that concepts such as antibiotics and one class of nephrotoxic drugs, non-steroidal anti-inflammatory drugs (NSAIDs), are significant for the prediction of Acute Kidney Injury in time series data. 

The adaptation of human-centered explanations in such domains where efficiently communicating with highly skilled experts is crucial showcases the potential of human-centered explanation in real-world problems.

\section{Conclusion and Discussion}

We have provided an overview of current advances in concept-based explanations to explain complex machine learning models. Concepts can be seen as a way to bridge the gap between the reasoning of  humans and machine learning models. Instead of having humans speak the language of machines, in terms of raw input features and training samples, concept-based explanations aim to speak the language of humans, via concepts. Concepts and their scores, as defined Sec. ~\ref{chihkuanyehsec:score}, can be used as quantitative tool for such translation. We have seen that concept-based methods can be combined with feature-based and sample-based explanations to leverage their strengths, and can also be used as a basis for contrastive explanations. The human-centered nature of concept explanations has also made concept explanation successful in several real-world applications, especially when highly skilled professionals with established domain concepts (e.g., medicine) are involved.

Despite this exciting progress, many interesting open questions remain. How can we discover unknown confounding concepts? Can we extend concept explanations to models such as tree models, where the assumption that concept vectors are directions in a linear vector space need no longer hold?
How can we identify the intrinsic properties of what makes a concept understandable to humans? For instance, in the high dimension activation space in the neural network, why are some vectors/directions understandable to human, while some others are not? Can we teach humans new concepts by identifying training examples that can be discriminated by directions in the activation space? More generally, can we use the richness of concepts to convey knowledge from a super-human-performance model to humans?

\bibliographystyle{plain}
\bibliography{reference}
\end{document}